\def\year{2021}\relax
\documentclass[letterpaper]{article} 
\usepackage{aaai21}  
\usepackage{times}  
\usepackage{helvet} 
\usepackage{courier}  
\usepackage[hyphens]{url}  
\usepackage{graphicx} 
\usepackage{natbib}  
\usepackage{caption} 
\usepackage{amsfonts}
\usepackage{diagbox}
\usepackage{color,xcolor}
\usepackage[ruled,vlined,linesnumbered]{algorithm2e}

\usepackage{soul}
\usepackage{url}
\usepackage{times}
\usepackage{helvet}
\usepackage{courier}
\usepackage{underscore}
\usepackage{graphicx}
\usepackage{textcomp}
\usepackage{xcolor}
\usepackage{bm}
\usepackage{amsmath,amssymb,amsfonts}
\usepackage{algpseudocode}     
\usepackage{times}
\usepackage{helvet}
\usepackage{courier}
\usepackage{verbatim}
\usepackage[ruled,vlined,linesnumbered]{algorithm2e}
\usepackage[separate-uncertainty]{siunitx}
\usepackage{multirow}
\usepackage{subfigure}

\def\genbox#1#2#3#4#5#6{
	\leavevmode\raise#4bp\hbox to#5bp{\vrule height#5bp depth0bp width0bp
		\pdfliteral{q .5 w \csname #2COLOR\endcsname\space RG
			\csname #3PDF\endcsname{#5}{#6} S Q
			\ifx1#1 q \csname #2COLOR\endcsname\space rg 
			\csname #3PDF\endcsname{#5}{#6} f Q\fi}\hss}}

\def\trianbox   #1#2{\genbox{#1}{#2}  {trian}    {0}   {5}    {2.5}}

\def\circbox    #1#2{\genbox{#1}{#2}  {circ}     {0}   {5}    {2.5}}

\begin{document}
	\title{Prototypical Networks for Multi-Label Learning}
	\author{Zhuo Yang$^1$, Yufei Han$^2$, Guoxian Yu$^3$, Qiang Yang$^1$, Xiangliang Zhang$^1$\\}
	\affiliations{
		$^1$King Abdullah University of Science and Technology, Thuwal, Saudi Arabia\\ 
		$^2$Norton Research Group, Sophia Antipolis, France\\ 
		$^3$Shandong University , Shandong Province, China\\ 
		$\{ $zhuo.yang, qiang.yang, xiangliang.zhang $\}$@kaust.edu.sa\\
		yfhan.hust@gmail.com\\
		gxyu@sdu.edu.cn\\
	}
	\maketitle
	
\begin{abstract}
We propose to formulate multi-label learning as a  estimation of class distribution  in a non-linear embedding space, where for each label,   its positive  data embeddings and negative data embeddings distribute compactly to form a positive component and negative component respectively, while the positive component and negative component are pushed away from each other. Duo to the shared embedding space for all labels, the distribution of embeddings preserves instances' label membership and feature matrix, thus encodes the feature-label relation and nonlinear label dependency. Labels of a given instance are inferred in the embedding space by measuring the probabilities of its belongingness to the positive or negative components of each label. Specially, the probabilities are modeled as the distance from the given instance to representative positive or negative prototypes. Extensive experiments validate that the proposed solution can provide distinctively more accurate  multi-label classification than other state-of-the-art algorithms.
\end{abstract}

\section{Introduction}

Multi-label learning addresses the problem that one instance  can be associated with multiple labels simultaneously. 
Formally, 
the goal is to learn a function $F$, which   maps an instance ${\bf{x}} \in \mathbb{R}^{D}$ to a label vector ${\bf{y}} =[ l_{1},l_{2}, \cdots , l_{K} ]$  ($l_{i}$ is 1 if ${\bf{x}}$ is associated with the $i$-th label, and $l_{i}$ is 0 otherwise).
Many real-world applications drive the study of this problem, such as image object recognition \cite{mlimage1,mlimage2}, text classification \cite{mltext1,mltext2},  and bioinformatic problems \cite{mlbio}.

How to exploit label dependency is the key to success in multi-label learning. Let ${\bf{X}} \in \mathbb{R}^{N\times D}$ denote all training instances, and ${\bf{Y}} \in \{0,1\}^{N \times K}$ be the corresponding label matrix. 
Previous research efforts study the label dependency mainly by 1) exploiting the label matrix $\textbf{Y}$ only \cite{jfsc,camel,embed1,embed3}; and 2) jointly mapping $\textbf{X}$ and $\textbf{Y}$ into a same low rank label space \cite{coembed1,coembed2,coembed3,mllrc}.
The approaches using solely $\textbf{Y}$ to extract label dependency only focus on how training instances carrying two different labels overlap with each other (which are the shared instances). They ignore the profiles of the shared instances (how they look like). 
Existing approaches using both $\textbf{X}$ and $\textbf{Y}$ often assume a low-rank linear structure of the label dependency. Simple and effective as it is, this assumption doesn't necessarily hold in real world applications and oversimplifies the underlying complex label correlation patterns \cite{sleec}.

\begin{figure}[t]
	\centerline{\includegraphics[width=0.85\columnwidth]{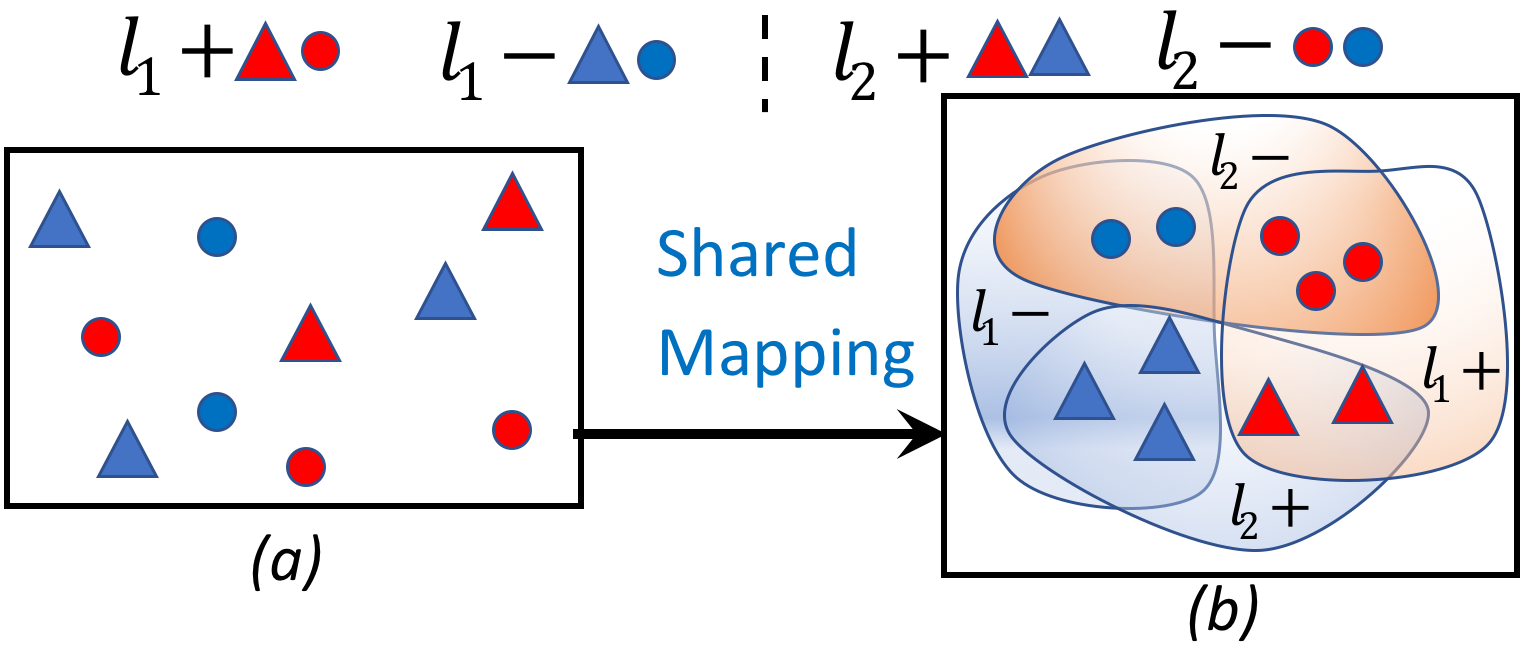}}
	\caption{Intuition of our study: non-separable instances in the original feature space \emph{(a)} are mapped to a non-linear embedding space \emph{(b)} via feature embedding process applied to all labels. Label $l_1$ is indicated by color (red for positive and blue for negative). Label $l_2$ is presented by shape (triangle for positive and circle for negative). In space \emph{(b)}, ${l_k} + $/${l_k} - $ indicates the positive/negative component of label $k$, and  data instances dropped into the area \trianbox1{cred} are tagged with both labels \{$l_1$,$l_2$\},  those in \circbox1{cred} or \trianbox1{cblue} only have \{$l_1$\} or \{$l_2$\}  and those in \circbox1{cblue} carry neither of them.  Label dependency is then captured by the distribution of instance embeddings in the new space.} 
	\label{dependency}
	\vspace{-0.6cm}
\end{figure}

Motivated by the previous research progress, we aim 
to exploit both $\textbf{X}$ and $\textbf{Y}$ to capture the non-linear label dependency. We attack the multi-label learning from a novel angle of distribution estimation. The intuition is that there exists an embedding space, in which for each label, its positive instances distribute compactly to form a positive component and the remained negative instances form a negative component, which is seperated as much as possible from the positive one. 
Figure \ref{dependency} demonstrates the intuition. Instances are mapped into the embedding space described by Figure \ref{dependency} \emph{(b)}, in which for each label $l_k (k=1,2)$,  its positive instances are grouped into component ${l_k} +$, which is seperated from the negative component ${l_k} -$. 
The key to capture label dependency  is the shared mapping function applied to all labels, without which (each label has its specific mapping function), classification for different labels  will be done in different spaces independently and no label dependency will be exploited. Specially, by shared mapping, one embedding in the new space can adjust its position by its label membership and feature profile, which jointly exploits information from matrices $\textbf{X}$, $\textbf{Y}$ and  encodes non-linear label dependency  into the distribution of embeddings.

We employ mixture density estimation  to model the distribution of a positive/negative component and depending on the complexity of  distributions, one positive/negative component can be represented by one or several clusters. 
We then define the mean of cluster as prototype and this comes the name of our approach \textbf{PNML} (\emph{Prototypical Networks for Multi-Label Learning}). In practice, our PNML is proposed to work under two modes, named PNML-multiple and PNML-single. With mode PNML-multiple, the optimal number and parameters of prototypes are learned based on an adaptive clustering-like process, which brings us strong representation power for components at the sacrifice of computation efficiency.  Mode PNML-single is thus proposed alternatively to compute one prototype for one component.
For a given instance, its class membership to one label can be measured by its Bregman distance to the positive/negative prototypes of this label. 
The distance metric  is learned for each label to model the different distribution pattern of each label.  
Specially, to exploit label dependency one more step, a designed label correlation regularizer is incorporated into the object function to carve these prototypes further.

We highlight our contributions  as follows:
\begin{enumerate}
	\item To the best of our knowledge, the proposed PNML method is the first attempt to address   multi-label learning   by \textbf{jointly estimating the class distribution of all labels in an nonlinear embedding space, which exploiting non-linear label dependency and feature-label predicative relation effectively.}
	\item \textbf{A Bregman divergence based distance metric is learned jointly for each label} alongside with the mixing distribution model of the prototypes. It measures the closeness in distribution from a data instance to the positive and negative components of each label.
	
	\item Extensive experiments including ablation study are conducted on 15   benchmark datasets. The results show the effectiveness of PNML, that PNML-multiple and PNML-single both achieve superior classification performance  to all baselines. 
	  
\end{enumerate}

\section{Related Work}
There have been many designs of various types of multi-label learning models. We concentrate on the discussion of the most recent and relevant work regarding the ways of exploiting label dependency.

Binary relevance based methods \cite{br} decompose a multi-label classification problem into $K$ independent  binary classification problems while ignoring label dependency. It is known as  \emph{the first-order approach}. In contrast, methods in \cite{jfsc,nop2} make use of the pair-wise label co-occurrence pattern, and are thus featured  as \emph{the second-order methods}. For example, in JFSC \cite{jfsc}, a pair-wise label correlation matrix ${\bf{C}}$ is calculated from $\bf{Y}$ and used to regularize the multi-label learning objective. 
RAKEL \cite{rakel} exploits higher-order label dependency from ${\bf{Y}}$ by grouping labels as mutually exclusive meta-labels,  so as to transform a multi-label  classification task as a multi-class problem w.r.t. the meta-labels. CAMEL \cite{camel} learns to represent any given label as a linear combination of all the labels of $\textbf{Y}$.
Among above approaches, JFSC \cite{jfsc} and CAMEL achieved state-of-the-art performance.
Other methods \cite{embed1,embed3,coembed1,coembed3,mllrc} exploit high-order label dependency by learning a low-rank representation of $\textbf{Y}$ and a predicative mapping function between $\textbf{X}$ and the low-rank label embedding. 
These two steps are conducted independently or jointly in these approaches.
Nevertheless, the main limit of these methods is the low-rank assumption of the label matrix, which doesn't necessarily hold in practices \cite{sleec}. Besides, the label dependency usually has a more complicated structure than the simple linear model. 

MT-LMNN \cite{dmetric2} and LM-kNN \cite{lmknn} also exploits both $\bf{X}$ and $\bf{Y}$ to extract the label dependency from the angle of metric learning. MT-LMNN treats the classification of each label as an individual task and a distance metric is learned for this label to keep an instance with this label stay closer to its neighbors also with this label.
LM-kNN maps one instance's feature profile and label vector onto a low dimensional manifold, where the projection of the feature and labels stay close to each other. Our approach can be also interpreted as a metric learning process, where an instance carrying or not a specific label $k$ is supposed to be close to the positive or negative prototypes of the label. 
Compared to MT-LMNN and LM-kNN, we jointly learn the prototypes and the distance metric of each label in the non-linear embedding space. Our approach is designed to be more flexible to capture the label dependency and the feature-label relation. 

Our method is inspired by prototypical networks \cite{prototype,prototype2}, which is originally designed for few-shot multi-class tasks.
While the prototypes of different classes in the multi-class problem are loosely correlated. 
In our study, 
the positive and negative prototypes of different labels are \emph{strongly correlated}.
Such label correlation is captured by learning the self-tuning mixture distribution of each label. Besides, we learn the Bregman divergence function jointly to measure the similarity between a data instance and the prototypes, which adapts better to the data distribution than the primitive Euclidean distance used in \cite{prototype,prototype2}. 


\section{Methodology}
\subsection{Overview of the Proposed Model}
\begin{figure*}[t]
	\centerline{\includegraphics[width=1.8\columnwidth]{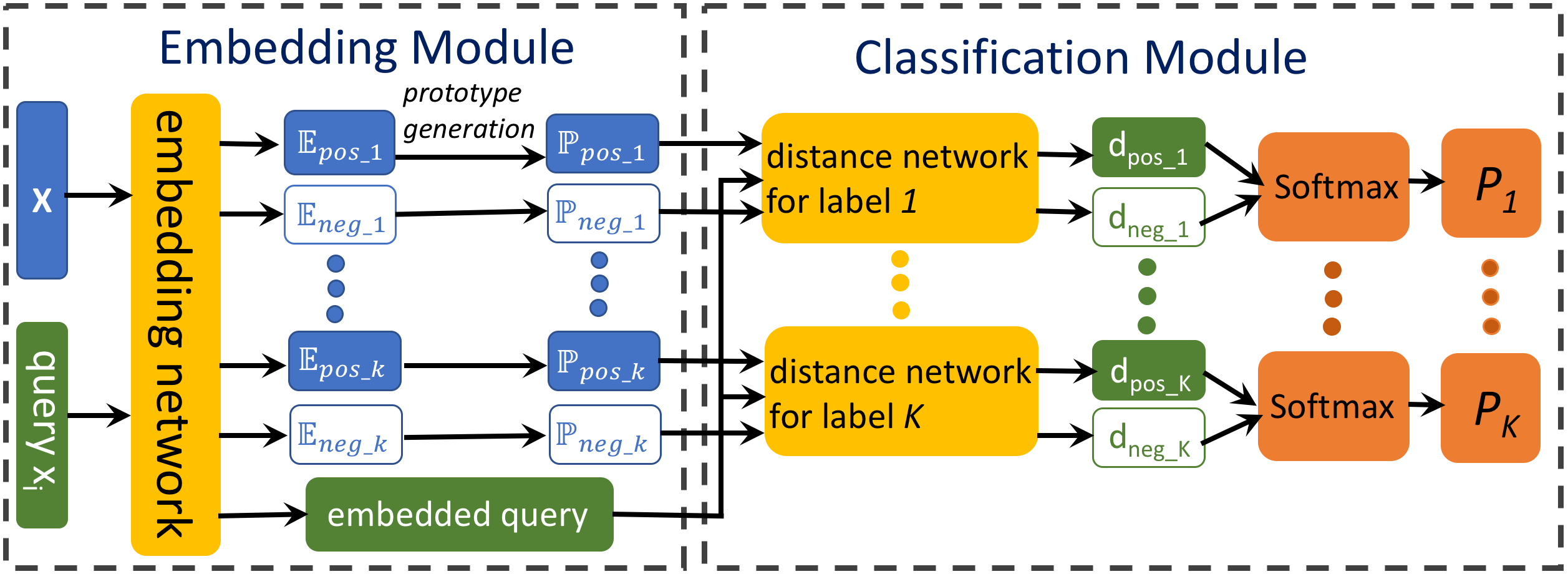}}
	\caption{Overview of the proposed model PNML. For $k$=$1,...,K$, ${{\mathbb{E}}}_{{pos}\_k}$ (${\mathbb{E}}_{{neg}\_k}$) is the set of embeddings of positive (negative) instances for label $k$, i.e., $\mathbb{E}_{{pos}\_k}=\{ {f_\phi }({\bf{x}}),{\bf{x}} \in {\mathbb{X}_{pos\_k}}\}$, where ${\mathbb{X}_{pos\_k}}$ is the positive instance set of label $k$. ${\mathbb{P}}_{{pos}\_k}$/${\mathbb{P}}_{{neg}\_k}$ is the positive/negative prototype set of label $k$. $d_{pos\_k}$/$d_{neg\_k}$ is the average distance from embedding of query ${{\bf{x}}_{\bf{i}}}$ to ${\mathbb{P}}_{{pos}\_k}$/${\mathbb{P}}_{{neg}\_k}$. $P_k$ is the predicted probability of ${{\bf{x}}_{\bf{i}}}$ having label $k$.
	}
	\label{framework}
	\vspace{-0.4cm}
\end{figure*}

Figure \ref{framework} shows the overall architecture of PNML. In the training stage, the {\bf embedding module} learns the non-linear feature embedding function applied to all the labels ${f_\phi }:{\mathbb{R} ^D} \to {\mathbb{R} ^M}$. Specially, this shared embedding function $f_\phi$ locates embedding in the new space by its label membership and original feature profile, which naturally encodes the information from label matrix ${\bf{Y}}$ and feature matrix ${\bf{X}}$ into the distribution of embeddings in the new space, and thus captures the non-linear label dependency.   
In Figure \ref{framework}, the \emph{Embedding Layer} network $f_\phi$ is defined by a one-layer fully connected neural network with LeakyRelu \cite{relu} as activation function. In the new embedding space, the distribution of data instances of each label $k$ is described as a mixture of local prototypes ${\mathbb{P}}_{{pos}\_k}$ and ${\mathbb{P}}_{{neg}\_k}$ of positive and negative components $\mathbb{E}_{{pos}\_k}$ and $\mathbb{E}_{{neg}\_k}$. $\mathbb{E}_{{pos}\_k}=\{ {f_\phi }({\bf{x}}),{\bf{x}} \in {\mathbb{X}_{pos\_k}}\}$, where ${\mathbb{X}_{pos\_k}}$ is the positive instance set of label $k$ and vice versa.  With respect to the different strategies of prototype generation, PNML is proposed to work under two different modes, mode \textbf{PNML-multiple} and mode \textbf{PNML-single}:
\begin{itemize}
	\item \textbf{PNML-multiple}: In this mode, prototypes in ${\mathbb{P}}_{{pos}\_k}$ and ${\mathbb{P}}_{{neg}\_k}$ are generated by an adaptive distance-based clustering-like process, in which the number and parameters of prototypes  are tuned jointly in the training process. Usually multiple prototypes are generated in this adaptive process, so we call this mode PNML-multiple.   
	\item \textbf{PNML-single}: In this mode,  one prototype for one   component is computed as the expectation of positive  or negative embeddings of each label, so we call this mode PNML-single.
\end{itemize}
\noindent Mode PNML-multiple prefers more precise representation of components $\mathbb{E}_{{pos}\_k}$ and $\mathbb{E}_{{neg}\_k}$ compared to mode PNML-single, while its adaptive prototype generation process pays much computation efficiency. Alternatively, PNML-single is reduced to compute one prototype for one component, which saves computation significantly without much classification accuracy loss. We will introduce the prototype generation methods in these two modes next part and compare the two modes' performance (classification accuracy and run time) in experiment part.
 
The {\bf classification module} learns a Bregman-divergence based distance function for every label, based on the prototypes ${\mathbb{P}}_{{pos}\_k}$ and ${\mathbb{P}}_{{neg}\_k}$.
In the testing stage, a query ${\bf x}_i$ is classified by going through the learned embedding layer to first have an embedding of ${\bf x}_i$. For each label, the distance between the embedding vector and each of the positive and negative prototypes is computed with the learned distance metric layer.
These distance measurements actually represent the probabilities of ${\bf x}_i$'s belongingness to each cluster (one cluster is described by one prototype).
\textit{softmax} is then performed on these probabilities to determine whether ${\bf x}_i$ carries the label. We elaborate the design of PNML in the followings. 

\subsection{Mixture Density Estimation}

Let ${\bf{e}}= f_\phi ({\bf x})$ be the embedding vector of   ${\bf x}$ in the embedding space. We assume that the class-conditional probability $p(\mathbf{e}|\hat{y}_{e}=\pm{1})$ of $\bf{e}$ belonging to positive or negative class of each label $k$ follows a mixture of the distributions of multiple positive or negative clusters, noted as $p(\left. {\bf{e}} \right|\pmb{\Omega}^{+/-} )$: 
\begin{equation}
\begin{array}{c}
p(\mathbf{e}|\hat{y}_{e}=\pm{1}) = p(\left. {\bf{e}} \right|\pmb{\Omega}^{+/-} ) = \sum\limits_{s=1}^{k_{+/-}} {{\pi^{+/-} _s}} {p_\psi }({\bf{e}}|{\pmb{\theta}^{+/-}_s})
\end{array}
\label{mixture}
\end{equation}
where $\Omega^{+/-} = \{\pi^{+/-}_{s},\pmb{\theta}^{+/-}_{s}\}$ are the learnable mixing coefficient and density function parameters of the mixture models of the positive and negative class. $k_{+/-}$ denotes the number of positive or negative clusters, which is tuned jointly in the training process in mode PNML-multiple and is set to 1 in mode PNML-single . For the convenience of analysis, we constrain each ${p_\psi }({\bf{e|}}\pmb{\theta}^{+/-}_{s} )$ to be an exponential family distribution function, with canonical parameters $\pmb{\theta}^{+/-}_{s}$:
\begin{equation}
\begin{array}{r}
{p_\psi }({\bf{e|}}\pmb{\theta}^{+/-}_{s} ) = h(\mathbf{e})\exp (T({{\bf{e}}})\pmb{\theta}^{+/-}_{s}  - \psi (\pmb{\theta}^{+/-}_{s} ))
\end{array}
\label{red}
\end{equation}
where $T(\mathbf{e})$ denotes sufficient statistics of the distribution of $\mathbf{e}$ and $\psi(\pmb{\theta}^{+/-}_{s})$ is the cumulant generating function, defined as the logarithm of the normalization factor to generate a proper probabilistic measure. $h(\mathbf{e})$ is the carrier measure.  

Given learned $\pmb{\Omega}^{+/-}$, we measure the posterior probability of $\bf{e}$ belonging to positive or negative class of each label as:
\begin{equation}
p({{\hat{y}}_{e} = +1 }|{\bf{e}}) = \frac{p(\left. {\bf{e}} \right|\pmb{\Omega}^{+} )}{p(\left. {\bf{e}} \right|\pmb{\Omega}^{+} ) + p(\left. {\bf{e}} \right|\pmb{\Omega}^{-} )}
\label{cluster}
\end{equation}
where the positive and negative class prior probabilities are set equally as $0.5$. Without specified prior domain knowledge, the non-informative prior is a reasonable choice. Eq. (\ref{cluster}) can be thus interpreted as a likelihood ratio test. 

According to the Theorem 4 in  \cite{mdensity}, there is a unique Bregman divergence associated with every member of the exponential family. For example, spherical Gaussian distribution is associated with  squared Euclidean distance
Therefore, we can rewrite the regular exponential family distribution given in Eq.(\ref{red}) by a regular Bregman divergence \cite{prototype,mdensity} as: 
\begin{equation}
\begin{array}{r}
{p_\psi }({\bf{e|}}\pmb{\theta}^{+/-}_{s} ) = \exp ( - {d_\varphi}({\bf{e}},\mu (\pmb{\theta}^{+/-}_{s})) - {g_\varphi}({\bf{e}}))
\end{array}
\label{bregman}
\end{equation}

\noindent where $d_\varphi $  is the unique Bregman divergence determined by the conjugate Legendre function of $\psi$. $g_{\psi}(\mathbf{e})$ absorbs all the rest terms that are not related to $\mathbf{e}$, but determined by $\psi$.  $\mu (\pmb{\theta} ) $ is the expectation of the exponential family distribution defined by Equation (\ref{expectation}). 
\begin{equation}
\begin{array}{r}
\mu (\pmb{\theta} ) = E_{p_\psi} [{\bf{e}}] =\int_{\mathbb{R} ^M}{{\bf{e}} {p_\psi }({\bf{e|}}\pmb{\theta} ) d {\bf{e}}}
\end{array}
\label{expectation}
\end{equation}

\noindent\textit{\textbf{Prototype Generation for Mode PNML-single.}} \ \ 
We define the expectation in Equation (\ref{expectation}) as prototype. In mode PNML-single, one positive/negative component of label $k$ is treated as one positive/negative cluster, so one positive/negative prototype is computed for it as:
\begin{equation}
{{\bf{P}}_{pos\_k/neg\_k}} = \frac{1}{{\left| {{{\mathbb{X}}_{pos\_k/neg\_k}}} \right|}}\sum\limits_{{{\bf{x}}_i} \in {{\mathbb{X}}_{pos\_k/neg\_k}}} {{f_\phi }({{\bf{x}}_i})} 
\end{equation}
where $\left| {{{\mathbb{X}}_{pos\_k/neg\_k}}} \right|$ is the size of set ${{{\mathbb{X}}_{pos\_k/neg\_k}}} $. Here  ${{\bf{P}}_{pos\_k}}$ composes the positive prototype set ${{\mathbb{P}}_{pos\_k}}$ and ${{\bf{P}}_{neg\_k}}$ composes the negative prototype set ${{\mathbb{P}}_{neg\_k}}$.

\noindent\textit{\textbf{Prototype Generation for Mode PNML-multiple.}} \ \ 
We adopt an adaptive process in mode PNML-multiple to generate prototype based on real-world data's statistical profiles.
In our study, without loss of generality, we concretize the distribution of each prototype described by Equation (\ref{red}). The Bregman divergence of each positive or negative prototype $d_{\psi}(\mathbf{e},\mu(\pmb{\theta}^{+/-}_{s}))$ is thus can be used to evaluate the probability of  $\mathbf{e}$'s belongingness to each prototype.
To learn these prototypes, we 
follow the idea of infinite mixture  distribution applied previously in \cite{dpmeans,prototype2}. The main steps are :
\begin{enumerate}
	\itemsep-0.1em
	\item Initialize ${\pmb{\mu}_c}$ =  $\mu_{\mathbb{E}_{{pos}\_k}}$ ($\mu_{\mathbb{E}_{{neg}\_k}}$) as the mean of ${\mathbb{E}}_{{pos}\_k}$ (${\mathbb{E}}_{{neg}\_k}$), $C=1$ as the initial number of prototypes, and ${\sigma _c} = \sigma$ 
	, which is the trainable variance of one cluster from which instance is assumed to be sampled. \emph{ite\_clustering} is the iteration number of clustering.
	\item Estimate the distance threshold (Eq.\ref{lamda}) for creating a new prototype:
	\begin{equation}
	\label{lamda}
	\lambda  =  - 2\sigma \log \left( {\frac{\alpha }{{{{(1 + \frac{\rho }{\sigma })}^{M/2}}}}} \right)
	\end{equation}
	where $\rho$ is a measure of the standard deviation for the base distribution from which prototypes are drawn, $M$ is the dimension of embedding vector and $\alpha$ is a hyperparameter named concentration parameter.
	\item For each embedding vector $\mathbf{e}_i$ in ${\mathbb{E}}_{{pos}\_k}$ (${\mathbb{E}}_{{neg}\_k}$), compute its' distance to each prototype $c$ in $\{ 1, \cdots ,C\} $ as ${d_{i,c}} = {d_\psi }({{\bf{e}}_{\bf{i}}}{\bf{,}}{\pmb{\mu} _{\bf{c}}}).$
	If ${\min _c}{d_{i,c}} > \lambda$ , set $C = C + 1$, update $\pmb{\mu}_c = {\mathbf{e}_i}$ and ${\sigma _c} = {\sigma}$. After that, compute the probability of $\mathbf{e}_i$ belonging to each cluster by ${z_{i,c}} = \frac{{\exp ( - {d_\varphi }({\mathbf{e}_i},{\pmb{\mu} _c}))}}{{\sum\nolimits_c {\exp ( - {d_\varphi }({\mathbf{e}_i},{\pmb{\mu} _c}))} }}$
	, and then recompute the cluster mean as ${\pmb{\mu} _c} = \frac{{\sum\nolimits_i {{z_{i,c}}} {\mathbf{e}_i}}}{{\sum\nolimits_i {{z_{i,c}}} }}$. 
	\item Repeat step 3   for \emph{ite\_clustering} times. Finally,  Each $\pmb{\mu}_c$ is a prototype vector and all $\pmb{\mu}_c$, $c=1,...,C$   compose   the prototype set ${\mathbb{P}}_{{pos}\_k}$ (${\mathbb{P}}_{{neg}\_k}$).
\end{enumerate}

With the defination of prototype, combining Equation (\ref{mixture}), (\ref{cluster}) and (\ref{bregman}) and adopting noninformative class and cluster prior probability, we can write the prediction probability of  query ${{\bf{x}}_{\bf{i}}}$'s having label $k$ as:
\begin{equation}
\resizebox{0.45\textwidth}{!}
{  
$p({\hat y_{i,k}} = + 1|{{\bf{x}}_i}) = \frac{{\frac{1}{{\left| {{{\mathbb{P}}_{pos\_k}}} \right|}}\sum\limits_{c = 1}^{\left| {{{\mathbb{P}}_{pos\_k}}} \right|} {\exp ( - {d_\varphi }({f_\phi }({{\bf{x}}_i}),{{\bf{P}}_{pos\_k\_c}}))} }}{{\sum\limits_{l \in \{ pos,neg\} } {\left[ {\frac{1}{{\left| {{{\mathbb{P}}_{l\_k}}} \right|}}\sum\limits_{c = 1}^{\left| {{{\mathbb{P}}_{l\_k}}} \right|} {\exp ( - {d_\varphi }({f_\phi }({{\bf{x}}_i}),{{\bf{P}}_{l\_k\_c}}))} } \right]} }}$
}
\label{prediction}
\end{equation}
where $\left| {{{\mathbb{P}}_{{pos\_k}/{neg\_k}}}} \right|$ is the size of positive/negative prototype set of label $k$. ${{\bf{P}}_{{pos\_k\_c}/{neg\_k\_c}}}$ is the $c$-th prototype vector in $ {{{\mathbb{P}}_{{pos\_k}/{neg\_k}}}}$.

Based on Equation \ref{prediction}, we can formulate the learning objective function of PNML as a cross-entropy loss:
\begin{equation}
\begin{array}{*{20}{c}}
{{L_e} =  - \sum\limits_{i = 1}^N {\sum\limits_{k = 1}^K {{y_{i,k}}\log p(} {{\hat y}_{i,k}} =  + 1|{{\bf{x}}_i})} }\\
{ + (1 - {y_{i,k}})\log (1 - p({{\hat y}_{i,k}} =  + 1|{{\bf{x}}_i}))}
\end{array}
\label{lossd}
\end{equation}

We next move to the second module of distance metric learning for classification.

\subsection{Label-Wise Distance Metric Learning}

In \cite{mdensity}, various distance functions $d_\varphi(\cdot,\cdot)$ are presented for popular distribution functions in exponential families. For example, multivariate  spherical Gaussian distribution ${p_\psi }({\bf e |\pmb{\theta}})$ is associated with  squared Euclidean distance $||{\bf{e}}-\mu (\pmb{\theta}) ||^2$. 
Although Euclidean distance is simple and showed its effectiveness in \cite{prototype}, it is not appropriate to our multi-label learning, because  components of our label distribution may not be spherical Gaussian distribution. 

The non-spherical distribution can be firstly caused by the instances' common membership among different categories. One component may be pulled to get close to several different components due to their common instances. Secondly, for multi-label learning, features may play different roles in different labels' discriminant processes \cite{lift,jfsc}, which implies that each label has its own specific  distribution pattern. 

We are  therefore encouraged to learn approximating the Bregman divergence $d_{\psi}(\mathbf{e},\pmb{\mu}(\theta_{s}))$ for each label by learning a Mahalanobis distance function ${d_m^k}$ given the mean vector of the prototypes in the embedding space, which shows: 
\begin{equation}
{d_m^k}({{\bf{e}} },{{ \pmb{\mu}_{s}} }) = \sqrt {({{\bf{e}} - {\pmb{\mu}_{s}})^T}{{\mathbf{U}_k}^T}{\mathbf{U}_k}({\bf{e}}  - {\pmb{\mu}}_{s})}
\label{maha}
\end{equation}
The non-spherical covariance structure is approximated by the weight matrix $\mathbf{U}_{k}\in{\mathbb{R}^{M\times{M}}}$ learned from the data \cite{dmetric1,dmetric2}. 

In this paper, we encode $\bf{U_k}$ with a one-layer fully connected neural network with a linear activation function, as shown in \emph{Distance Network} in Figure \ref{framework}. 
Eq. (\ref{lossd}). 
Moreover, we add a regularizer described by Equation (\ref{lossl}) to the overall loss function to prevent over-fitting of the distance metric learning.
\begin{equation}
{L_m} = \sum_{k=1}^{K}\left\| {\bf{U}_k} \right\|_F^2
\label{lossl}
\end{equation}

\subsection{Label Correlation Regularizer}\label{lcr}
In our approach, the prototypes can be viewed as labels' representatives, especially the positive prototypes. In practice, the  negative instances for a label are usually much more than the  corresponding positive instances. Then  for many labels, their negative instances are mostly in common. Consequently the negative prototypes of labels may be similar, and less interesting than positive prototypes.
We thus take the positive prototypes and enhance the correlation between positive prototypes of different labels in the training process.
Intuitively, if label $j$ and $k$ are positively correlated in $\bf{Y}$, the mean of their positive prototypes should be close as well and thus present a large inner product. 
We introduce an regularization term defined by:
\begin{equation}
{L_c} = \frac{1}{2}\sum\limits_{j = 1}^K {\sum\limits_{k = 1}^K {(1 - {c_{jk}})\mu {{({{\mathbb{P}}_{pos\_j}})}^T}\mu ({{\mathbb{P}}_{pos\_k}})} } 
\label{lossc}
\end{equation}
where $c_{jk}$ indicates the correlation measurement between $j$-th column and $k$-th column of label matrix $\bf Y$. ${\mu ({{\mathbb{P}}_{pos\_j}})}$ and ${\mu ({{\mathbb{P}}_{pos\_k}})}$ are the mean vector of prototype set ${{{\mathbb{P}}_{pos\_j}}}$ and ${{{\mathbb{P}}_{pos\_k}}}$.

The overall loss function is given in Eq. (\ref{wloss}), where
$\lambda _1$ and $\lambda _2$ are the non-negative tradeoff parameters.
\begin{equation}
{L_{all}} = {L_e} + {\lambda _1}{L_m} + {\lambda _2}{L_c}
\label{wloss}
\end{equation}

\subsection{Training procedure}
In the training process of our approach, for each label, we need to load the feature matrix $\bf{X}$ for mapping and building prototypes. In mode PNML-single, the mean of embedding set $\mathbb{E}_{{pos}\_k}$ and $\mathbb{E}_{{neg}\_k}$ needs to be computed during every training iteration, and in mode PNML-multiple, distance from every instance embedding ${{\bf{e}}_i}$ to cluster centroids needs to be compared to threshold. The computational cost significantly increases when  data sets get larger. To mitigate this issue, we can sample instances for each label to reduce the amount of data involved for prototype computing.  We denote ${r_{pos\_k}}$ and ${r_{neg\_k}}$ as the sampling rate for positive and negative instances, respectively. 
Usually, compared to ${r_{neg\_k}}$, ${r_{pos\_k}}$ is set bigger to sample as many positive instances as possible since they are more informative and rare, while ${r_{neg\_k}}$ is set smaller to reduce computation significantly without losing much information. The influence of the sampling rate on model performance will be evaluated in next section. 

The training procedure first samples positive and negative instances respectively. Then the network weights for embedding ${f_\phi }$, distance metric $d_m^k(\cdot, \cdot)$ and the prototypes' parameters are updated. Adam \cite{adam} is used as the optimizer. 

\section{Experiments}
We organised comprehensive experiments in this section. Due to the limited space, we intorducte the implementation platforms in supplementary document, also further experimental results. 
\subsection{Experimental Setup}
\noindent \textit{\textbf{Datasets and Evaluation Metrics.}} \ \ Fifteen public benchmark datasets are used to evaluate all the involved approaches comprehensively. 
These datasets have different application contexts, including text, biology, music and image. 
Table \ref{dataset} summarizes the details of these data sets. 
We choose $5$ popularly applied evaluation metrics to measure the performances \cite{metrics}. They are $\textit{{Accuracy}}$, \textit{Macro-averaging $F_{1}$}, \textit{Micro-averaging $F_{1}$}, $\textit{{Average\ \ precision}}$ and $\textit{{Ranking\ \ Loss}}$.
\vspace{-0.1cm}
\begin{table}[!htb]
	\scriptsize                   
	\small                    
	\caption {Used   benchmark multi-label datasets. $N$/$D$ denotes the number of instances/features of a data set. $Labels$ denotes the number of label in dataset.  $Card$ denotes the average number of labels associated with each instance.}
	\vspace{-0.2cm}
	\label{dataset}
	\renewcommand\arraystretch{0.7}
	\newcommand{\tabincell}[2]{\begin{tabular}{@{}#1@{}}#2\end{tabular}}
	\centering
	\begin{tabular}{c|c|c|c|c|c} 
		\hline
		\hline
		dataset	& $N$ & $D$ & $Labels$  &$Card$  &domain \\
		\hline
		emotions	&593 	&72  &6  &1.869	 & music		\\
		scene	&2407 	&294  &5  &1.074	 & image		\\
		image	&2000 	&294  &5  &1.240	 & image		\\
		arts	&5000 	&462  &26  &1.636	 & text(web)	\\
		science &5000   &743  &40  &1.451    & text(web)   \\
		education &5000   &550  &33  &1.461    & text(web)   \\
		enron	&1702 	&1001  &53  &3.378	 & text	\\
		genbase	&662 &1186  &27  &1.252	 & biology		\\
		rcv1-s1	&6000 	&944  &101  &2.880	 & text	\\
		rcv1-s3	&6000 	&944  &101  &2.614	 & text	\\
		rcv1-s5	&6000 	&944  &101  &2.642	 & text	\\
		bibtex	&7395 	&1836  &159  &2.402	 & text	\\	
		corel5k	&5000 	&499  &374  &3.522	 & image	\\	
		bookmark & 87856 &2150&208&2.028& text \\
		imdb & 120919&1001&28&2.000&text \\	
		\hline
		\hline
	\end{tabular}
	\vspace{-0.3cm}
\end{table}

\noindent\textit{\textbf{Comparing Approaches.}} \ \ 
We compare PNML with the following 6 multi-label learning approaches, including first-order, second-order and high-order approaches. Some of these approaches achieve state-of-the-art multi-label classification performance and half of them were proposed within 3 years. Specially, an state-of-the-art extreme multi-lable classification approach is also introduced for broder comparison. 
The comparing approaches are:
\begin{itemize}
	\item \textit{BR} \cite{br}: It decomposes multi-label problem as $K$ independent single-label classification problems and is a first-order approach 
	\item \textit{ML-KNN} \cite{mlknn}: It is derived from traditional KNN method for multi-label problem and it is a high-order approach.
	\item \textit{MLTSVM} \cite{mltsvm}: It is adapted from SVM.
	\item \textit{JFSC} \cite{jfsc}: It does feature selection and classification for each label jointly. It is a second-order approach and is one of the state-of-the-art approach.
	\item \textit{CAMEL} \cite{camel}: It treats each label as a linear combination of all other labels. It is a high-order approach and is one of the state-of-the-art approach.
	\item \textit{Parabel} \cite{parabel}:It is one of the   state-of-the-art  multi-label extreme classification methods. Since \textit{Parabel} is designed specially for ranking, so to keep fairness, we just compare their performance in terms of the ranking metrics, that is $\textit{{Ranking\ \ Loss}}$ and $\textit{{Average\ \ precision}}$. 
\end{itemize}

Baseline models of \textit{BR}, \textit{ML-KNN} and \textit{MLTSVM} are implemented with the scikit-learn package \cite{skp} and a two-layer multi-layer perceptron is used as the base classifier for \textit{BR}. The unit number of hidden layer of MLP is determined by Eq.(\ref{dimension}) where $D$ is the input feature dimension. Besides, the number of nearest neighbors of \textit{ML-KNN} is searched in $\{ 3,5, \cdots , 21\}$.
For \textit{MLTSVM}, the empirical risk penalty parameter $c_k$ and regularization parameter ${\lambda _k}$ are searched in $\{ {2^{ - 6}},{2^{ - 5}},{2^{ - 4}}, \cdots ,{2^{ 4}}\} $. Codes and suggested parameters in the original papers are used for \textit{JFSC}, \textit{CAMEL} and \textit{Parabel}. 

In our approach, the embedding dimension $M$, slope of LeakyRelu $\beta$, concentration parameter $\alpha$ (only needed by PNML-multiple mode) and loss tradeoff parameters $\lambda _1$, $\lambda _2$ are hyperparameters to be determined. Empirically, $\beta$ is set to $0.2$ and $M$ is determined by Eq.(\ref{dimension}), in which $D$ is the feature dimension in original feature space. $\lambda _1$ and $\lambda _2$, they are searched in $\{ {10^{ - 7}},5 \times {10^{ - 6}},{10^{ - 6}},5 \times {10^{ - 6}}, \cdots ,{10^{ - 2}}\} $. When under mode PNML-multiple, $\alpha$ is searched in
$\{0.0001,0.001, 0.01, 0.1, 0.5, 1.0\}$.
\begin{equation}
M = \left\{ {\begin{array}{*{20}{c}}
	{72}\\
	{128}
	\end{array}\begin{array}{*{20}{c}}
	{ D \le 200}\\
	{D > 200}
	\end{array}} \right.
\label{dimension}
\end{equation}

\subsection{Classification Results}\label{results}
\begin{table*}[!htb]
	\small                    
	\setlength\tabcolsep{4pt} 
	\caption{Experimental results of evaluated algorithms on 15 data sets on 5 evaluation metrics.
		$ \uparrow \left(  \downarrow  \right)$ indicates the larger (smaller) the value, the better the performance. The best results are  in bold . \emph{AR} is the average rank of algorithm on 15 data sets with corresponding metric.}
	\vspace{-0.5cm}
	\label{result1}
	\begin{center}
		\renewcommand\arraystretch{0.8}
		\resizebox{\textwidth}{!}{ 
			\begin{tabular}{c|c c c c c c c c c c c c c c c|c}
				\hline
				\hline
				\textbf{Comparing}&\multicolumn{16}{|c}{\textbf{$Accuracy$} $ \uparrow $}  \\
				\cline{2-17} 
				\textbf{Algorithm} & \textbf{emotions}& \textbf{scene}& \textbf{image} & \textbf{arts}& \textbf{science}& \textbf{education}& \textbf{enron} & \textbf{genbase}& \textbf{rcv1-s1}& \textbf{rcv1-s3}& \textbf{rcv1-s5}& \textbf{bibtex}&\textbf{corel5k}& \textbf{bookmark}& \textbf{imdb}& \emph{AR}\\
				\hline
				BR      & 0.478 & 0.672 & 0.530 & 0.341 & 0.350 & 0.365 & 0.419  & 0.988  & 0.324 & 0.378 & 0.381 & 0.321 & 0.106  &0.233&0.065& 4.47\\
				ML-KNN      & 0.418 & 0.692 & 0.522 & 0.189 & 0.280 & 0.324 & 0.273  & 0.973  & 0.338 & 0.342 & 0.349 & 0.313& 0.079 & 0.242&0.145& 5.27 \\
				MLTSVM      & 0.320 & 0.418 & 0.500 & 0.241 & 0.309 & 0.283 & 0.273  & 0.952  & 0.235 & 0.245 & 0.312 & 0.201& 0.091  & 0.173&0.071&6.47 \\				
				JFSC      & 0.255 & 0.603 & 0.451 & 0.377 & 0.365 & 0.346 & 0.418  & \textbf{0.994}  & 0.354 & 0.391 & 0.397 & 0.352 & \textbf{0.142} & 0.250& \textbf{0.248}&3.60 \\
				CAMEL      & {0.505} & 0.702 & {0.589} & 0.342 & 0.331 & 0.360 & 0.440  & 0.992  & 0.300 & 0.354 & 0.358 & 0.312  & 0.079 & 0.239&0.098&4.20\\
				Parabel     & -- & -- & -- & -- & -- & -- & -- & -- & -- & -- & -- & -- & -- & -- & -- & --  \\
				PNML-single     & {0.489} & {0.708} & {0.583} & {0.381} & {0.394} & {0.394} & {0.472}   & 0.987  & {0.395} & {0.422} & {0.428} & {0.387} & 0.121 &0.253 &0.109&2.53\\		
				PNML-multiple     & \textbf{0.509} & \textbf{0.728} & \textbf{0.603} & \textbf{0.393} & \textbf{0.402} & \textbf{0.400} & \textbf{0.475}  & 0.990  & \textbf{0.398} & \textbf{0.427} & \textbf{0.431} & \textbf{0.390} & {0.123} & \textbf{0.258} &0.121&1.33\\			
				\hline			
				
				\hline
				\textbf{Comparing}&\multicolumn{16}{|c}{\textbf{$Micro-averaging\ \ F_{1}$} $  \uparrow $} \\
				\cline{2-17} 
				\textbf{Algorithm} & \textbf{emotions}& \textbf{scene}& \textbf{image} & \textbf{arts}& \textbf{science}& \textbf{education}& \textbf{enron} & \textbf{genbase}& \textbf{rcv1-s1}& \textbf{rcv1-s3}& \textbf{rcv1-s5}& \textbf{bibtex}&\textbf{corel5k}& \textbf{bookmark}& \textbf{imdb}& \emph{AR}\\
				\hline
				BR      & 0.609 & 0.735 & 0.615 & 0.400 & 0.410 & 0.442 & 0.540  & 0.990  & 0.426 & 0.435 & 0.449 & 0.426 & 0.185  &0.244&0.105 &4.73\\
				ML-KNN      & 0.548 & 0.748 & 0.592 & 0.250 & 0.348 & 0.403 & 0.425 & 0.979  & 0.426 & 0.394 & 0.413 & 0.416  & 0.143 &\textbf{0.317} &0.180&5.33 \\
				MLTSVM      & 0.484 & 0.559 & 0.560 & 0.319 & 0.376 & 0.376 & 0.436 & 0.963  & 0.379 & 0.362 & 0.453 & 0.340 & 0.160  & 0.226&0.137&6.27 \\	
				JFSC      & 0.407 & 0.697 & 0.552 & {0.444} & 0.446 & 0.445 & 0.542 & \textbf{0.994}  & 0.495 & 0.487 & 0.497 & 0.471 & \textbf{0.243}  & 0.281 &\textbf{0.342}&3.33\\
				CAMEL      & {0.636} & \textbf{0.770} & {0.659} & 0.409 & 0.421 & 0.450 & 0.564  & 0.993  & 0.403 & 0.413 & 0.431 & 0.421  & 0.110 & 0.278 &0.166&3.87\\
				Parabel     & -- & -- & -- & -- & -- & -- & -- & -- & -- & -- & -- & -- & -- & -- & -- & --  \\
				PNML-single     & 0.630 & 0.744 & 0.648 & 0.440 & {0.454} & {0.471} & {0.596}   & 0.988  & {0.533} & {0.516} & {0.532} & {0.501} & 0.198& 0.262 &0.194&2.87\\
				PNML-multiple     & \textbf{0.648} & {0.757} & \textbf{0.662} & \textbf{0.446} & \textbf{0.462} & \textbf{0.474} & \textbf{0.598}   & 0.991  & \textbf{0.534} & \textbf{0.518} & \textbf{0.540} & \textbf{0.502} & 0.207& {0.268}&0.203&1.53\\	
				\hline
				
				\hline
				\textbf{Comparing}&\multicolumn{16}{|c}{\textbf{$Macro-averaging\ \ F_{1}$} $ \uparrow $} \\
				\cline{2-17} 
				\textbf{Algorithm} & \textbf{emotions}& \textbf{scene}& \textbf{image} & \textbf{arts}& \textbf{science}& \textbf{education}& \textbf{enron} & \textbf{genbase}& \textbf{rcv1-s1}& \textbf{rcv1-s3}& \textbf{rcv1-s5}& \textbf{bibtex}&\textbf{corel5k}& \textbf{bookmark}& \textbf{imdb}& \emph{AR}\\
				\hline
				BR      & 0.578 & 0.746 & 0.612 & 0.245 & 0.241 & 0.221 & 0.231  & 0.747  & 0.271 & 0.244 & 0.244 & 0.311  & 0.041 & 0.173&0.049&3.47\\				
				ML-KNN      & 0.528 & 0.755 & 0.589 & 0.146 & 0.169 & 0.176 & 0.140  & 0.674  & 0.239 & 0.184 & 0.208 & 0.267 & 0.027 & 0.191&0.040&5.33\\
				MLTSVM      & 0.479 & 0.570 & 0.564 & 0.200 & 0.173 & 0.156 & 0.127  & 0.600  & 0.201 & 0.162 & 0.135 & 0.205 & 0.036   & 0.104&0.032&6.40\\	
				JFSC      & 0.352 & 0.701 & 0.546 & 0.235 & 0.222 & 0.144 & 0.196  & 0.746  & 0.256 & 0.225 & 0.232 & 0.354 & 0.039 & 0.136&0.080&4.80\\
				CAMEL      & 0.619 & \textbf{0.779} & {0.662} & 0.232 & 0.206 & 0.198 & 0.212  & \textbf{0.755}  & 0.177 & 0.156 & 0.156 & 0.251 & 0.026 & 0.139 &0.087&4.40\\
				Parabel     & -- & -- & -- & -- & -- & -- & -- & -- & -- & -- & -- & -- & -- & -- & -- & --  \\
				PNML-single      & {0.634} & 0.756 & 0.655 & \textbf{0.328} & {0.296} & \textbf{0.310} & {0.258}   & 0.746  & 0.378 & {0.365} & {0.372} & \textbf{0.418} & 0.064 & \textbf{0.232}&\textbf{0.156}&1.87\\
				PNML-multiple     & \textbf{0.647} & {0.767} & \textbf{0.666} & {0.321} & \textbf{0.298} & {0.304} & \textbf{0.262}   & 0.733  & \textbf{0.389} & \textbf{0.375} & \textbf{0.377} & {0.414} & \textbf{0.066}& 0.228&0.144&1.67\\	
				\hline
				
				\hline
				\textbf{Comparing}&\multicolumn{16}{|c}{\textbf{$Average\ \ precision$} $ \uparrow $} \\
				\cline{2-17} 
				\textbf{Algorithm} & \textbf{emotions}& \textbf{scene}& \textbf{image} & \textbf{arts}& \textbf{science}& \textbf{education}& \textbf{enron} & \textbf{genbase}& \textbf{rcv1-s1}& \textbf{rcv1-s3}& \textbf{rcv1-s5}& \textbf{bibtex}&\textbf{corel5k}& \textbf{bookmark}& \textbf{imdb}& \emph{AR}\\
				\hline
				BR      & 0.604 & 0.721 & 0.632 & 0.378 & 0.369 & 0.388 & 0.418  & 0.989  & 0.314 & 0.376 & 0.377 & 0.318  & 0.276 & 0.424&\textbf{0.504}&7.00\\
				ML-KNN      & 0.652 & 0.824 & 0.682 & 0.380 & 0.410 & 0.453 & 0.521  & 0.982  & 0.537 & 0.541 & 0.558 & 0.348 & 0.218& 0.299&0.399&6.93\\	
				MLTSVM      & 0.636 & 0.800 & 0.770 & 0.545 & 0.560 & 0.592 & 0.680  & 0.997  & 0.569 & 0.605 & 0.585 & 0.520 & 0.241   & 0.401&0.424&5.60\\				
				JFSC      & 0.720 & 0.851 & 0.788 & {0.604} & 0.592 & 0.626 & 0.641  & {0.997}  & 0.600 & 0.619 & 0.630 & 0.594 & 0.292& 0.441&0.496&3.87\\
				CAMEL      & {0.783} & \textbf{0.893} & {0.825} & 0.596 & 0.604 & 0.621 & 0.678   & {0.997}  & 0.604 & {0.633} & 0.631 & 0.606 & \textbf{0.293}& 0.463&0.479&2.80\\
				Parabel     & 0.613 & 0.861 & 0.794 & 0.349 & 0.481 & 0.474 & 0.479  & \textbf{0.998}  & 0.589 & 0.616 & 0.635 & 0.604 & 0.267 & 0.458 &0.483&4.87\\
				PNML-single     & 0.769 & 0.867 & 0.816 & {0.609} & {0.608} & {0.635} & \textbf{0.682}  & 0.992  & {0.627} & 0.630 & {0.640} & {0.608}  & 0.273&0.476 &0.445&2.93\\
				PNML-multiple     & \textbf{0.788} & {0.872} & \textbf{0.828} & \textbf{0.622} & \textbf{0.611} & \textbf{0.638} & \textbf{0.682}   & 0.994 &\textbf {0.633} & \textbf{0.635} & \textbf{0.644} & \textbf{0.611} & 0.286& \textbf{0.481}&0.450&1.73\\	
				\hline
				
				\hline
				\textbf{Comparing}&\multicolumn{16}{|c}{\textbf{$Ranking\ \ loss$} $ \downarrow $} \\
				\cline{2-17} 
				\textbf{Algorithm} & \textbf{emotions}& \textbf{scene}& \textbf{image} & \textbf{arts}& \textbf{science}& \textbf{education}& \textbf{enron} & \textbf{genbase}& \textbf{rcv1-s1}& \textbf{rcv1-s3}& \textbf{rcv1-s5}& \textbf{bibtex}&\textbf{corel5k}& \textbf{bookmark}& \textbf{imdb}& \emph{AR}\\
				\hline
				BR      & 0.490 & 0.289 & 0.437 & 0.611 & 0.591 & 0.587 & 0.477   & 0.010  & 0.583 & 0.546 & 0.535 & 0.616  & 0.153& 0.095&\textbf{0.159}&6.93\\				
				ML-KNN      & 0.308 & 0.101 & 0.257 & 0.192 & 0.135 & 0.104 & 0.111  & 0.012  & 0.067 & 0.067 & 0.061 & 0.160  & 0.134& 0.187&0.218&5.73\\
				MLTSVM      & 0.374 & 0.102 & 0.175 & 0.128 & 0.099 & 0.077 & \textbf{0.077}  & 0.004  & 0.043 & \textbf{0.040} & 0.060 & 0.065 & 0.143   & 0.172&0.204&4.00\\					
				JFSC      & 0.242 & 0.090 & 0.176 & 0.162 & 0.151 & 0.104 & 0.130   & 0.002  & 0.065 & 0.065 & 0.064 & 0.081 & 0.178& 0.146&0.167&4.80\\
				CAMEL      & {0.183} & \textbf{0.065} & {0.150} & 0.170 & 0.140 & 0.135 & 0.103  & 0.003  & 0.073 & 0.069 & 0.063 & 0.095  & 0.225& 0.101&0.189&4.87\\
				Parabel     & 0.356 & 0.075 & 0.172 & 0.272 & 0.171 & 0.168 & 0.208  & \textbf{0.001}  & 0.068 & 0.065 & 0.061 & 0.084 & 0.312 & 0.132  &0.163&4.93\\
				PNML-single      & 0.192 & 0.077 & {0.150} & {0.118} & {0.101} & {0.072} & {0.079}   & 0.002  & {0.037} & {0.043} & {0.039} & {0.071} & 0.143& 0.092&0.174&2.93\\
				PNML-multiple     & \textbf{0.181} & {0.076} & \textbf{0.147} & \textbf{0.110} & \textbf{0.093} & \textbf{0.070} & \textbf{0.077}   & \textbf{0.001}  & \textbf{0.035} & {0.041} & \textbf{0.038} & \textbf{0.062} & \textbf{0.131}& \textbf{0.088}&0.171&1.40\\	
				\hline
				\hline
			\end{tabular}
		}
	\end{center}
	\vspace{-0.4cm}
\end{table*}
\begin{table}[!htb]
	\scriptsize                   
	\caption {Results of pairwise comparison applied to PNML-single (PNML-multiple) with baseline algorithms.}
	\vspace{-0.2cm}
	\label{results2}
	\renewcommand\arraystretch{0.7}
	\newcommand{\tabincell}[2]{\begin{tabular}{@{}#1@{}}#2\end{tabular}}
	\centering
	\begin{tabular}{c|c|c|c|c|c} 
		\hline
		\hline
		\tabincell{c}{comparing \\ algorithms} & win & tie & lose  &$\begin{array}{c}
		V/2 + 1.96\sqrt V /2
		\end{array}$  &superior \\
		\hline
		BR	&73 	&0  &2  &\multirow{7}{*}{45.987 ($V=75$)}	 & yes		\\
		ML-KNN	&73 	&0  &2  &	 & yes		\\
		MLTSVM	&72 	&1  &2  &	 & yes		\\
		
		JFSC	& 62	&0  &13  &	 & yes		\\
		CAMEL	&59 	&0  &16  & 	 & yes	\\
		\hline
		Parabel & 24 &1 & 5  & 20.368 ($V=30$) & yes \\
		
		\hline
		\hline
	\end{tabular}
	\vspace{-0.4cm}
\end{table}

For each comparing approach, 5-fold cross-validation is performed on the training data of each data set. Tables \ref{result1} reports the average results of each comparing algorithm over 15 data sets in terms of each evaluation metric. Table \ref{results2} summarizes
the overall pairwise comparison results by comparing PNML-single (PNML-multiple) with  other baselines.
Based on these experimental results, the following observations can be made.
\begin{itemize}
	\item Our approach \textit{PNML-single (PNML-multiple)} outperforms the baseline approaches in most cases. Concretely, if we treat one evaluation metric for one dataset as one case, there are 75 cases in total.  \textit{PNML-single} outperforms all the other baselines in 73\% ($55/75$) evaluation cases, and \textit{PNML-multiple} outperforms them in 73\% ($55/75$) evaluation cases. Besides, the average rank of \textit{PNML-single} and \textit{PNML-multiple} is higher than all the other baselines with every evaluation metric. 
	Moreover, in Table \ref{results2}, pairwise comparison is done between  \textit{PNML-single (PNML-multiple)} and other baseline algorithms. 
	Generally, we have $15 \times 5 = 75$ cases, while for \emph{Parabel}, we have $15 \times 2 = 30$ cases.
	The sign test \cite{signtest} is employed to test whether \textit{PNML-single (PNML-multiple)} achieves a competitive performance against the other comparing algorithms. If the number of wins is at least $V/2 + 1.96\sqrt V /2$ ($V$ is the number of cases), the algorithm is significantly better with significance level $\alpha  < 0.05$. The results of indicate \textit{PNML-single (PNML-multiple)} is significantly superior to other baselines.
	\item In terms of evaluation metric, \textit{PNML-single (PNML-multiple)} performs best on \textit{Macro-averaging $F_{1}$}, which indicates \textit{PNML-single (PNML-multiple)} is more friendly to rarely encountered labels comparing to other approaches. This observation matches the results of \cite{prototype}, in which prototypical networks are used to solve few-shot learning problem. 
	\item \textit{PNML-multiple} outperforms \textit{PNML-single} in most cases, which indicates the learned multiple prototypes can describe the embedding distribution more comprehensively than one prototype. While \textit{PNML-single} is more efficient than \textit{PNML-multiple}. See supplementary document for detailed run-time evaluation.

\end{itemize}

\subsection{Ablation Study}

Without loss of generality, we tune the architecture of our proposed approach to verify the effectiveness of different parts under mode PNML-multiple. 
\begin{itemize}
	\item To confirm that our approach can learn well the label dependency and perform an effective mapping ${f_\phi }$, we assign different independently trained embedding layers to each label. Therefore $K$ single-label classifications are conducted independently without interaction. We name this variant of the proposed approach as {\bf \emph{PNML-I}}.
	\item  To demonstrate the effectiveness of the distance metric learning component, we build another variant of our approach in which Euclidean distance is used for each label and we name it
	{\bf \emph{PNML-D}}.
\end{itemize}
Table \ref{ablation2} summarizes the overall pairwise comparison results by comparing \textit{PNML-multiple} with \textit{PNML-I} and  \textit{PNML-D}.
The superiored results verifies that the learned embedding function ${f_\phi }$ does transfer meaningful knowledge across labels, the learned label-specific distance metric achieves better distribution modelling than Euclidean. See supplementary document for detailed results.
\begin{table}[!htb]
	\scriptsize                   
	\caption {Results of pairwise comparison applied to PNML-multiple with PNML variants.}
	\vspace{-0.2cm}
	\label{ablation2}
	\renewcommand\arraystretch{0.7}
	\newcommand{\tabincell}[2]{\begin{tabular}{@{}#1@{}}#2\end{tabular}}
	\centering
	\begin{tabular}{c|c|c|c|c|c} 
		\hline
		\hline
		\tabincell{c}{comparing \\ algorithms} & win & tie & lose  &$\begin{array}{c}
		V/2 + 1.96\sqrt V /2
		\end{array}$  &superior \\
		\hline
		
		PNML-I &68   &1  &6   &  \multirow{2}{*}{45.987 ($V=75$)} & yes   \\
		PNML-D &75   &0  &0   &  & yes   \\
		
		\hline
		\hline
	\end{tabular}
	\vspace{-0.7cm}
\end{table}

\subsection{Influence of Instance Sampling Rate}
\begin{figure}[!htb]
	\centerline{\includegraphics[width=0.9\columnwidth,height=4cm]{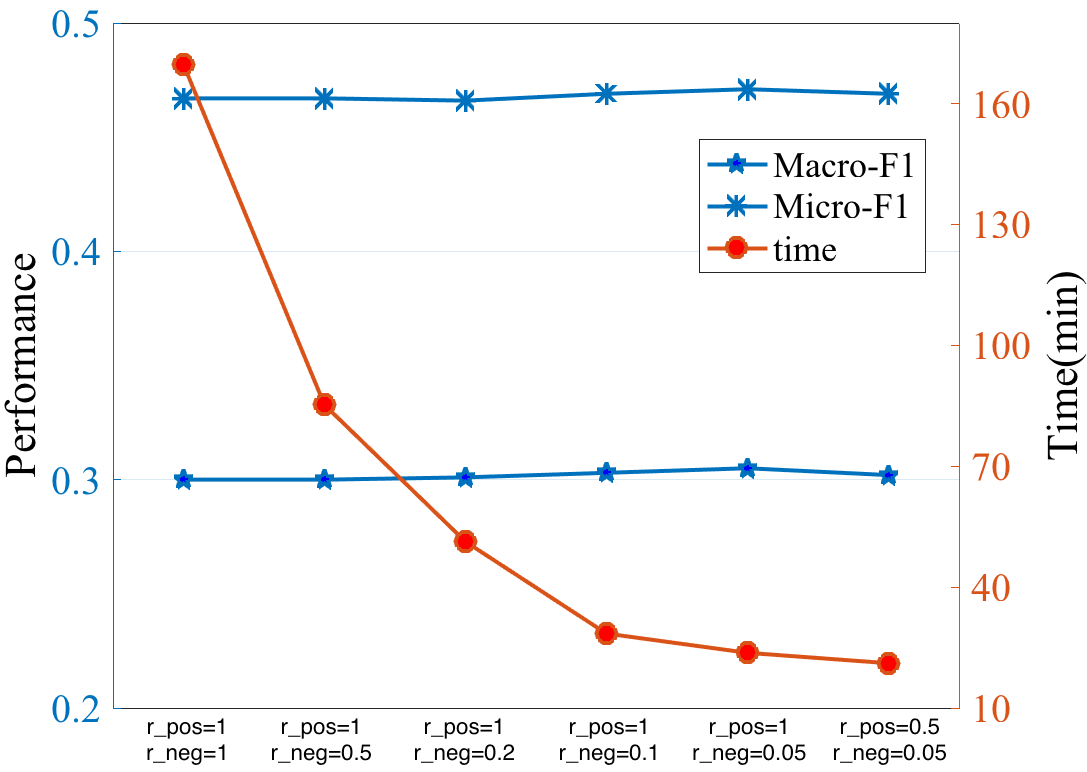}}
	\caption{Performance and run-time under different sampling rates.}
	\label{sample}
	\vspace{-0.6cm}
\end{figure} 
In our approach, to reduce the computational cost, we sample positive instances and negative instances for each label at rate  $r_{pos\_k}$ and $r_{neg\_k}$ in the training process. Here, we  show the influence of sampling rates on model predictive performance and efficiency. We choose dataset \textit{arts} for the experimental study and record the change of \textit{Macro-averaging $F_{1}$}, \textit{Micro-averaging $F_{1}$} and run-time of 5 folds under different sampling rates in Figure \ref{sample}. Here PNML works under the mode PNML-single. It can be observed that the performance keeps nearly the same under different sampling rates, while  run-time drops down with the decreasing of sampling rate. 

\section{Conclusion and Future Work}
In this paper, we propose PNML to
addresse multi-label learning by  estimation of class distribution in a new embedding space. While assuming each label has a mixture distribution of two components, a  mapping function  is learned to map label-wise training instances to two compact clusters, one for each component. Then positive and negative prototypes are defined for each label based on the distribution of embeddings (mode PNML-multiple) or as the expectations of positive embeddings and negative embeddings (mode PNML-single). 
Our approach can be extended to weak label problem and positive unlabeled problem in the future.

\clearpage


\bibliography{pnml.bib}

\clearpage

\appendix

\section{Further Experimental Results}

\subsection{Ablation Study}
Table \ref{ablation} gives the detailed results of ablation study.
\begin{table*}[!htb]
	\small                    
	\setlength\tabcolsep{4pt} 
	\caption{Experimental results of evaluated algorithms on 15 data sets on 5 evaluation metrics.
		$ \uparrow \left(  \downarrow  \right)$ indicates the larger (smaller) the value, the better the performance. The best results are  in bold . \emph{AR} is the average rank of algorithm on 15 data sets with corresponding metric.}
	\label{ablation}
	\begin{center}
		\renewcommand\arraystretch{1.0}
		\resizebox{\textwidth}{!}{ 
			\begin{tabular}{c|c c c c c c c c c c c c c c c}
				\hline
				\hline
				\textbf{Comparing}&\multicolumn{15}{|c}{\textbf{$Accuracy$} $ \uparrow $}  \\
				\cline{2-16} 
				\textbf{Algorithm} & \textbf{emotions}& \textbf{scene}& \textbf{image} & \textbf{arts}& \textbf{science}& \textbf{education}& \textbf{enron} & \textbf{genbase}& \textbf{rcv1-s1}& \textbf{rcv1-s3}& \textbf{rcv1-s5}& \textbf{bibtex}&\textbf{corel5k}& \textbf{bookmark}& \textbf{imdb}\\
				\hline
				PNML-I     & {0.492} & 0.697 & 0.572 & 0.364 & 0.351 & 0.385 & 0.420  & 0.987  & 0.352 & 0.384 & 0.397 & 0.356 & 0.121  & 0.221&0.108\\
				PNML-D      & 0.415 & 0.669 & 0.556 & 0.361 & 0.366 & 0.386 & 0.430  & 0.974  & 0.366 & 0.387 & 0.389 & 0.307  & 0.109 & 0.198&0.089\\	
				PNML-multiple     & \textbf{0.519} & \textbf{0.728} & \textbf{0.603} & \textbf{0.393} & \textbf{0.402} & \textbf{0.400} & \textbf{0.475}   & \textbf{0.990}  & \textbf{0.398} & \textbf{0.427} & \textbf{0.431} & \textbf{0.390} & \textbf{0.123} & \textbf{0.258}&\textbf{0.121}\\			
				\hline			
				
				\hline
				\textbf{Comparing}&\multicolumn{15}{|c}{\textbf{$Micro-averaging\ \ F_{1}$} $  \uparrow $} \\
				\cline{2-16} 
				\textbf{Algorithm} & \textbf{emotions}& \textbf{scene}& \textbf{image} & \textbf{arts}& \textbf{science}& \textbf{education}& \textbf{enron} & \textbf{genbase}& \textbf{rcv1-s1}& \textbf{rcv1-s3}& \textbf{rcv1-s5}& \textbf{bibtex}&\textbf{corel5k}& \textbf{bookmark}& \textbf{imdb}\\
				\hline
				PNML-I      & 0.628 & 0.751 & 0.642 & 0.417 & 0.407 & 0.450 & 0.540  & 0.989  & 0.470 & 0.453 & 0.475 & 0.452  & 0.190 & 0.239&0.191\\
				PNML-D      & 0.552 & 0.711 & 0.623 & 0.421 & 0.425 & 0.460 & 0.556  & 0.974  & 0.501 & 0.478 & 0.499 & 0.411 & 0.174 & 0.214&0.183\\
				PNML-multiple     & \textbf{0.653} & \textbf{0.757} & \textbf{0.662} & \textbf{0.446} & \textbf{0.462} & \textbf{0.474} & \textbf{0.598}  & \textbf{0.991}  & \textbf{0.534} & \textbf{0.518} & \textbf{0.540} & \textbf{0.502} & \textbf{0.207}& \textbf{0.268}&\textbf{0.203}\\	
				\hline
				
				\hline
				\textbf{Comparing}&\multicolumn{15}{|c}{\textbf{$Macro-averaging\ \ F_{1}$} $ \uparrow $} \\
				\cline{2-16} 
				\textbf{Algorithm} & \textbf{emotions}& \textbf{scene}& \textbf{image} & \textbf{arts}& \textbf{science}& \textbf{education}& \textbf{enron} & \textbf{genbase}& \textbf{rcv1-s1}& \textbf{rcv1-s3}& \textbf{rcv1-s5}& \textbf{bibtex}&\textbf{corel5k}& \textbf{bookmark}& \textbf{imdb}\\
				\hline
				PNML-I      & 0.641 & 0.762 & 0.642 & 0.288 & 0.240 & 0.255 & 0.242  & \textbf{0.737}  & 0.343 & 0.280 & 0.294 & 0.348 & 0.061 & 0.224&\textbf{0.146}\\
				PNML-D     & 0.554 & 0.739 & 0.633 & 0.313 & 0.282 & 0.280 & 0.253  & 0.719  & 0.367 & 0.353 & 0.354 & 0.375  & 0.058& 0.209&0.137\\
				PNML-multiple     & \textbf{0.652} &\textbf{0.767} & \textbf{0.666} & \textbf{0.328} & \textbf{0.298} & \textbf{0.311} & \textbf{0.262}   & 0.733  & \textbf{0.389} & \textbf{0.375} & \textbf{0.377} & \textbf{0.420} & \textbf{0.066}& \textbf{0.228}&0.144\\	
				\hline
				
				\hline
				\textbf{Comparing}&\multicolumn{15}{|c}{\textbf{$Average\ \ precision$} $ \uparrow $} \\
				\cline{2-16} 
				\textbf{Algorithm} & \textbf{emotions}& \textbf{scene}& \textbf{image} & \textbf{arts}& \textbf{science}& \textbf{education}& \textbf{enron} & \textbf{genbase}& \textbf{rcv1-s1}& \textbf{rcv1-s3}& \textbf{rcv1-s5}& \textbf{bibtex}&\textbf{corel5k}& \textbf{bookmark}& \textbf{imdb}\\
				\hline
				PNML-I      & 0.791 & 0.867 & 0.815 & 0.601 & 0.571 & 0.620 & 0.630  & \textbf{0.996}  & 0.580 & 0.591 & 0.594 & 0.544  & 0.271& 0.432&0.389\\
				PNML-D      & 0.693 & 0.860 & 0.807 & 0.601 & 0.599 & 0.627 & 0.646   & 0.989  & 0.593 & 0.604 & 0.606 & 0.577 & 0.260& 0.417&0.384\\
				PNML-multiple     & \textbf{0.795} &\textbf {0.872} & \textbf{0.828} & \textbf{0.622} & \textbf{0.611} & \textbf{0.638} & \textbf{0.682}   & 0.994 &\textbf {0.633} & \textbf{0.635} & \textbf{0.644} & \textbf{0.611} & \textbf{0.286}& \textbf{0.481}&\textbf{0.450}\\	
				\hline
				
				\hline
				\textbf{Comparing}&\multicolumn{15}{|c}{\textbf{$Ranking\ \ loss$} $ \downarrow $} \\
				\cline{2-16} 
				\textbf{Algorithm} & \textbf{emotions}& \textbf{scene}& \textbf{image} & \textbf{arts}& \textbf{science}& \textbf{education}& \textbf{enron} & \textbf{genbase}& \textbf{rcv1-s1}& \textbf{rcv1-s3}& \textbf{rcv1-s5}& \textbf{bibtex}&\textbf{corel5k}& \textbf{bookmark}& \textbf{imdb}\\
				\hline
				PNML-I      & 0.183 & \textbf{0.075} & 0.156 & 0.118 & 0.112 & 0.081 & 0.110   & \textbf{0.001}  & 0.056 & 0.063 & 0.057 & 0.090 & 0.158& 0.113&0.184\\
				PNML-D      & 0.281 & 0.080 & 0.160 & 0.117 & 0.100 & {0.072} & {0.079}  & 0.003  & 0.040 & 0.044 & 0.045 & 0.070   & 0.164& 0.127&0.193\\
				PNML-multiple     & \textbf{0.171} & {0.076} & \textbf{0.147} & \textbf{0.110} & \textbf{0.093} & \textbf{0.070} & \textbf{0.077}   & \textbf{0.001}  & \textbf{0.035} & \textbf{0.041} & \textbf{0.038} & \textbf{0.062} & \textbf{0.131}& \textbf{0.088}&\textbf{0.171}\\	
				\hline
				\hline
			\end{tabular}
		}
	\end{center}
\end{table*}

\subsection{Run Time Evaluation}
In this part, we show the run time evaluation experiments of PNML.  Table \ref{runtime} lists the one-fold training time of mode PNML-single and PNML-multiple on representative datasets under chosen sampling rates. 
The experiments were conducted on a Linux system using Python and our method is implemented using the Keras library. Each
experiment was conducted on an Nvidia 1080TI GPU. Batch size was set as 128 and epoch number was set as 40 for all datasets. We can obverse that PNML can finish training within hour even for large-scale dataset, especially mode PNML-single.  Besides, $t_{single}$ is much smaller than $t_{multiple}$, indicating that the adaptive prototype generation process dominates  the training complexity.
In part \ref{results}, we show that using adaptive prototype generation improves the classification performance in terms of all the used five evaluation metrics, comparing to the usage of a single prototype. So ,for  high-accuracy demanding applications, PNML-multiple can be adopted,     
and for  applications with efficiency requirement, PNML-single performs also well, better than the previous existing approaches.

\begin{table}[!htb]
	\small                    
	\setlength\tabcolsep{1.5pt} 
	\caption { 
		Run-time evaluation on representative datasets.  $N$/$D$ denotes the number of instances/features of a dataset. $Labels$ denotes the  number of labels in a dataset. $r_{pos\_k}$ and $r_{neg\_k}$ are the sampling rates. $t_{single}$(h) is the total training time of mode PNML-single in hours.  $t_{multiple}$ is the total training time of mode PNML-multiple. }
	
	\label{runtime}
	\newcommand{\tabincell}[2]{\begin{tabular}{@{}#1@{}}#2\end{tabular}}
	\centering
	\begin{tabular}{c|c|c|c|c|c|c} 
		\hline
		\hline
		dataset	& $N$ & $D$ & $Labels$ &$r_{pos\_k}$,$r_{neg\_k}$ &$t_{single}$(h)   & $t_{multiple}$(h) \\
		\hline
		emotions	&593 	&72  &6 &1, 0.5 & 0.001  & 0.006 	 		\\
		image	&2000 	&294  &5  & 1, 0.1 & 0.003 	& 	0.017  \\
		arts	&5000 	&462  &26  & 0.5, 0.05& 0.071 	& 0.597	\\
		enron	&1702 	&1001  &53  & 1, 0.1&0.029 	& 0.233	\\
		yeast	&2417 	&103  &14  & 1, 0.1&0.008 	&0.063	\\
		genbase	&662 &1186  &27  & 1, 0.5&0.002 	 &0.010	\\
		medical	&978 	&1449  &45  & 1, 0.5&0.004 	&0.028	\\
		rcv1-s1	&6000 	&944  &101  & 0.5, 0.05& 0.356 	& 3.685	\\
		bibtex	&7395 	&1836  &159  & 0.5, 0.05&0.735 	& 8.649	\\	
		corel5k	&5000 	&499  &374  & 0.5, 0.05&0.944 	& 10.576 \\		
		\hline
		\hline
	\end{tabular}
\end{table}

\subsection{Prototypes Visualization and Label Dependency}
\begin{figure}[!htb]
	\centerline{\includegraphics[width=0.9\columnwidth]{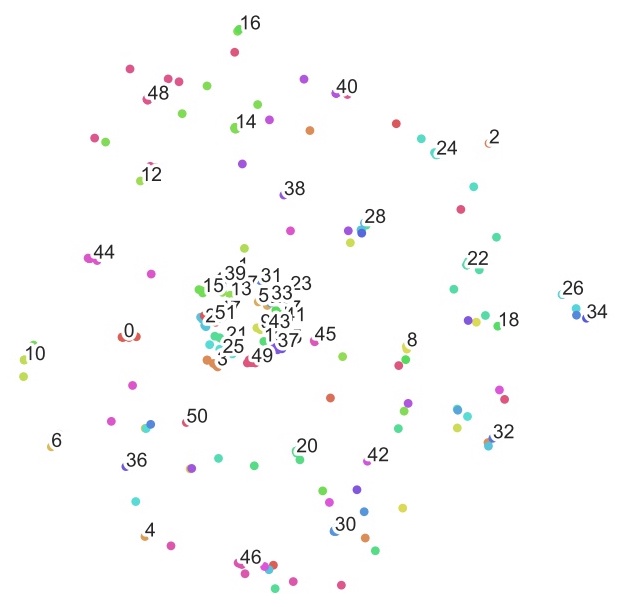}}
	\caption{tSNE visualization of prototypes learned on dataset \emph{arts} under mode PNML-multiple. Each point corresponds to one prototype, and prototypes belonging to the same positive or negative component of one label have the same color. Odd number locates at the mean of one label's negative prototypes. Even number locates at the mean of one label's positive prototypes. For example, number 0 indicates positive prototype mean of label 0, and number 1 indicates negative prototype mean of label 0.}
	\label{prototype}
\end{figure}

In our approach PNML, positive and negative prototypes are learned for positive and negative components of each label respectively. Based on our idea, positive and negative prototypes of one label should be pushed away from each other. Besides, in section \ref{lcr}, positive prototypes are adopted to build label correlation regularizer and the negative ones are proposed to be similar and less informative. Here we visualize the learned prototypes of dataset \textit{arts} under mode PNML-multiple in Figure \ref{prototype} to verify these points. From Figure \ref{prototype}, we can observe that negative prototypes (with odd numbers) stay together, implying that they are similar, and positive prototypes (with even numbers) are pushed away from negative ones. Though the similar are these negative prototypes, they play an important role as opposite of positives ones, which eases the classification process.

\begin{figure}[!htb]
	\centerline{\includegraphics[width=0.9\columnwidth]{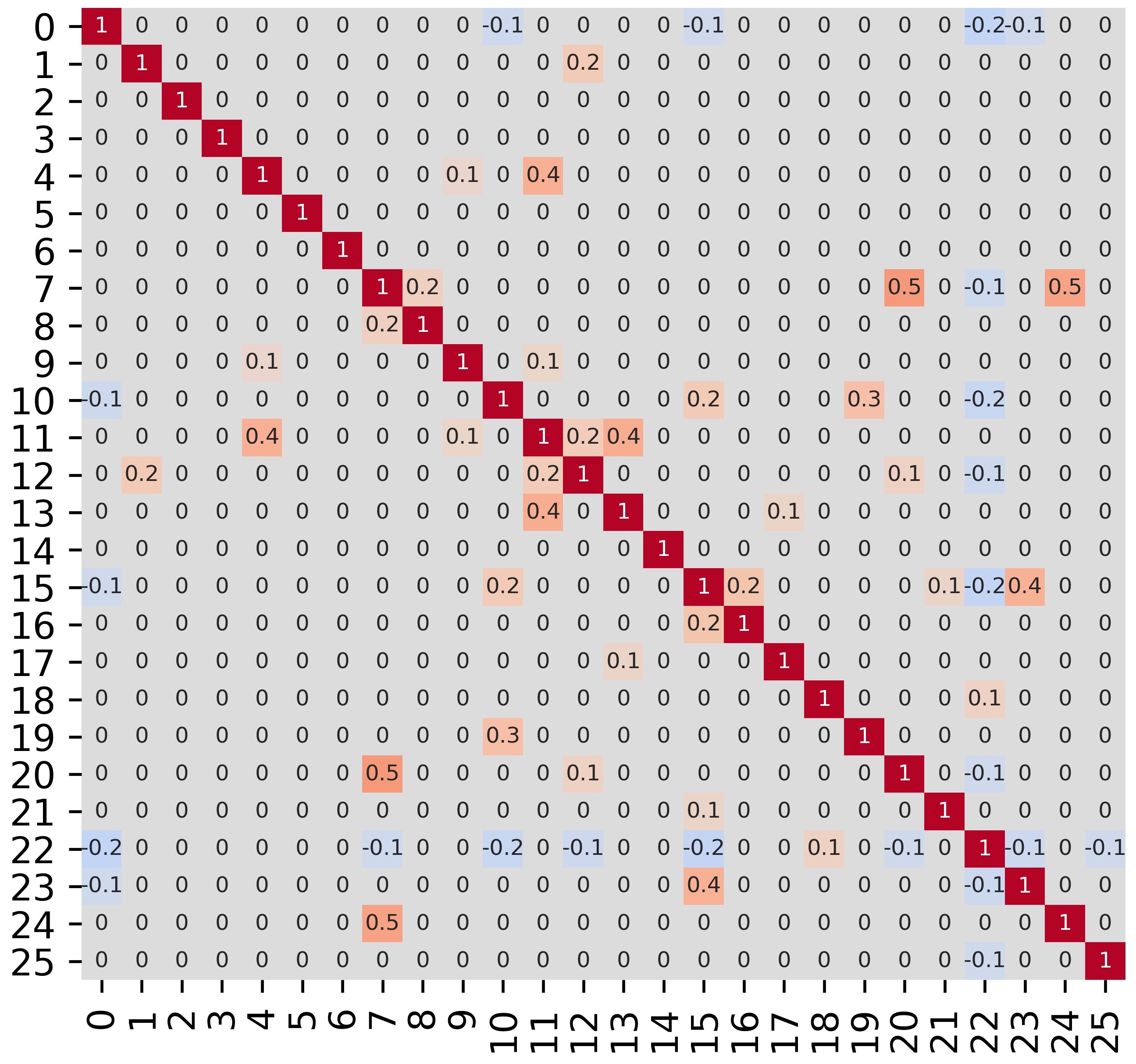}}
	\caption{Correlation coefficients computed by label matrix of dataset \textit{arts}. Correlation coefficients $coef < 0.1$ and $coef > -0.1$ are set to zero}
	\label{denl}
\end{figure}

Besides, in our PNML,  label dependency is encoded into prototypes, especially positive prototypes. Intuitively, if label $j$ and label $k$ have high positive label dependency, they should have many common positive instances, and thus similar positive prototypes. Figure \ref{denl} shows the roughly estimated label dependency represented by correlation coefficients computed by label matrix columns. In Figure \ref{denl}, label correlation coefficients of dataset \textit{arts} is showed and for the clear view of highly correlated labels, correlation coefficients $coef < 0.1$ and $coef > -0.1$ are set to 0. Compare Figure \ref{prototype} with Figure \ref{denl}, we can observe that prototypes present the label dependency. For example, in Figure \ref{denl}, label 7 has high  correlation coefficients with label 20 and label 24, and you can observe in Figure \ref{prototype} that the positive prototypes of these three labels (located by number 14, 40 and 48) stay close. An negatively correlatted example is the label 15 and label 22. You can observe that they have correlation coefficient $-0.2$ and the positive prototypes of them (located by number 30 and 44) are far away from each other.

\subsection{Parameter Sensitivity }
\begin{figure*}[!htb]	
	\centering
	\subfigure[$\lambda_1$]{
		\begin{minipage}[t]{0.3\linewidth}
			\centering{\includegraphics[width=1.0\columnwidth,height=3.5cm]{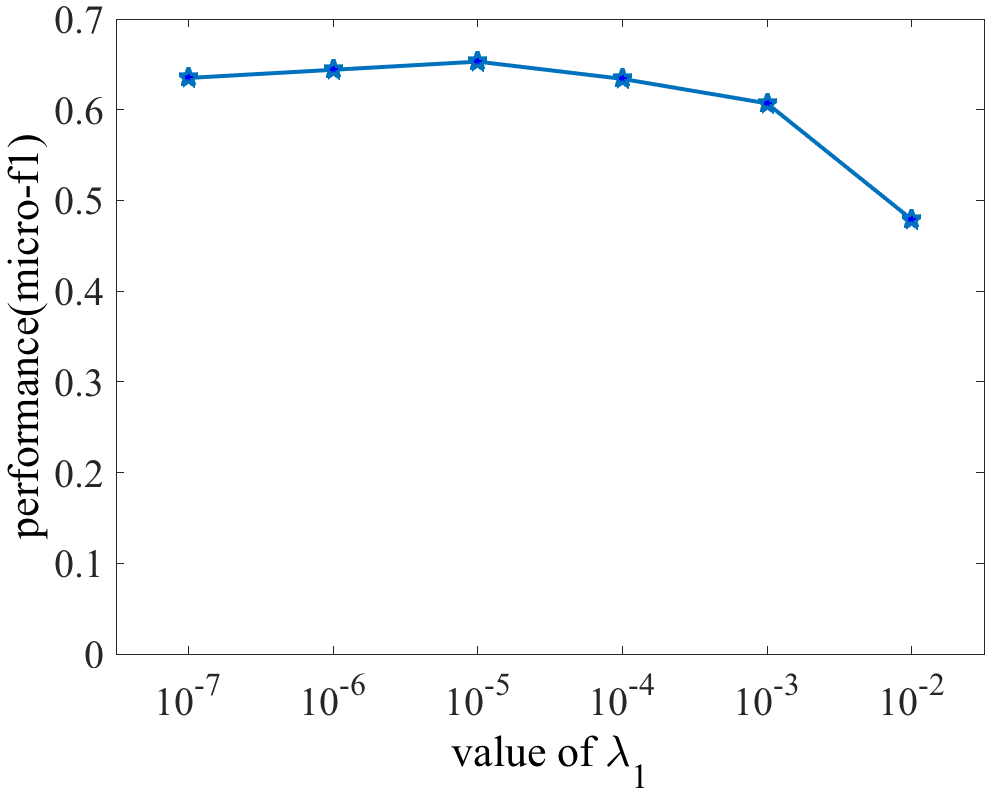}}			
		\end{minipage}
	}
	\subfigure[$\lambda_2$]{
		\begin{minipage}[t]{0.3\linewidth}
			\centering{\includegraphics[width=1.0\columnwidth,height=3.5cm]{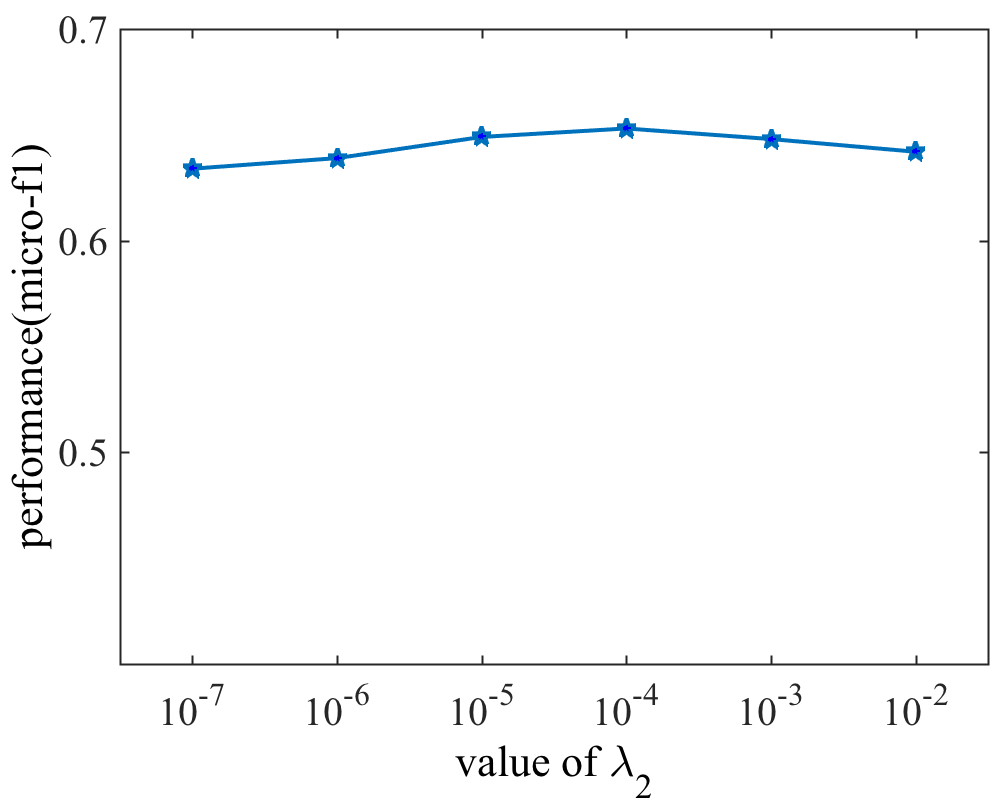}}			
		\end{minipage}
	}
	\subfigure[$\alpha$]{
		\begin{minipage}[t]{0.3\linewidth}
			\centering{\includegraphics[width=1.0\columnwidth,height=3.5cm]{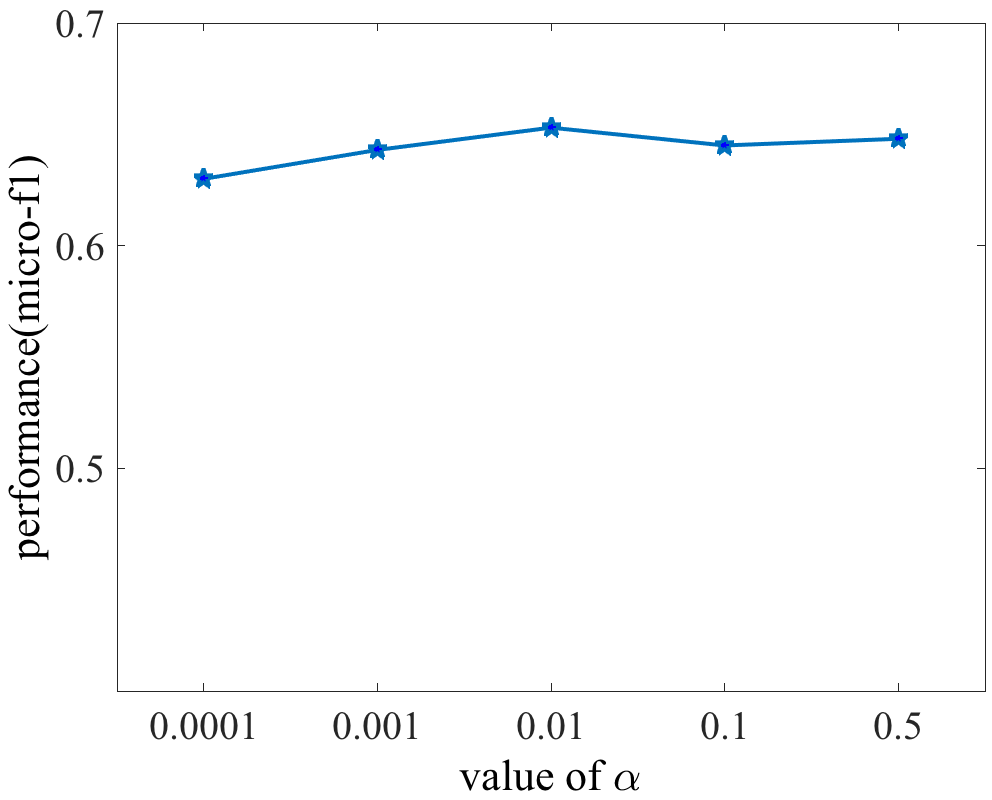}}			
		\end{minipage}
	}
	\centering
	\vspace{-0.3cm}
	\caption{ Parameter  sensitivity on dataset \emph{emotions} under mode PNML-multiple}
	\label{ps}
	\vspace{-0.6cm}
\end{figure*}

In this part, we show the sensitivity of parameters.
There are four/five hyperparameters in mode PNML-single/PNML-multiple, $M$ for embedding dimension, $\beta$ for slope of LeakyRelu, $\alpha$ for concentration parameter to determine the distance threshold (only used in mode PNML-multiple) and  $\lambda_1$, $\lambda_2$ for loss tradeoff parameters. $\beta$ is setted as $0.2$ and $M$ is setted by Equation \ref{dimension} in our paper. These two parameters are empirically setted and work well for all datasets we used. For $\lambda_1$, $\lambda_2$ and $\alpha$, Figure \ref{ps} shows their sensitivity test on dataset \emph{emotions} with metric \textit{Micro-averaging $F_{1}$} under mode PNML-multiple. The sensitivity test results for $\lambda_1$ and $\lambda_2$ under mode PNML-single will not be shown here since the similar trends.

\end{document}